\documentclass[letterpaper, 10 pt, conference]{ieeeconf}

\IEEEoverridecommandlockouts
\usepackage{graphicx}
\usepackage{amsmath, amssymb, amsfonts}
\usepackage{mathtools}
\usepackage{bm}
\usepackage{circuitikz}
\usepackage{tabularx}
\usepackage{multirow}
\usepackage{color, xcolor}
\usepackage{cite}
\usepackage{verbatim}
\usepackage{mathrsfs}
\usepackage{subcaption}  
\usepackage{svg}
\usepackage{empheq}
\usepackage{lipsum}
\usepackage{comment}
\usepackage{siunitx}
\usepackage[ruled,vlined]{algorithm2e}
\usepackage[english]{babel}
\usepackage{hyperref}
\usepackage{soul}  

\graphicspath{{./figures/}} 

\newcommand{\myfigure}[3]{
    \begin{figure}[t]
        \centering
        \includegraphics[width=0.925\linewidth]{#1}
        \caption{{#2}}
        \label{#3}
    \end{figure}
}

\newcommand{\myfigureSmaller}[3]{
    \begin{figure}[t]
        \centering
        \includegraphics[width=0.625\linewidth]{#1}
        \caption{{#2}}
        \label{#3}
    \end{figure} 
}

\newcommand{\myfigureMedium}[3]{
    \begin{figure}[t]
        \centering
        \includegraphics[width=0.8\linewidth]{#1}
        \caption{{#2}}
        \label{#3}
    \end{figure} 
}

\title{\LARGE \bf 
Robotic 3D Flower Pose Estimation for Small-Scale Urban Farms}

\author{Harsh Muriki$^{1}$, Hong Ray Teo$^{2}$, Ved Sengupta$^{1}$, and Ai-Ping Hu$^{3}$
\thanks{$^{1}$Georgia Institute of Technology, Atlanta, GA USA
{\tt\small (vmuriki3@gatech.edu, vsengupta7@gatech.edu)}.
 }
\thanks{$^{2}$Cornell University, Ithaca, NY USA
{\tt\small ht526@cornell.edu}.
 }
 \thanks{$^{3}$Georgia Tech Research Institute, Atlanta, GA USA
{\tt\small ai-ping.hu@gtri.gatech.edu}.
 }}

\begin{document}

\maketitle
\thispagestyle{empty}
\pagestyle{empty}

\begin{abstract}
The small scale of urban farms and the commercial availability of low-cost robots (such as the FarmBot) that automate simple tending tasks enable an accessible platform for plant phenotyping. We have used a FarmBot with a custom camera end-effector to estimate strawberry
plant flower pose (for robotic pollination) from acquired 3D point cloud models.
We describe a novel algorithm that translates individual occupancy grids 
along orthogonal axes of a point cloud
to obtain 2D images corresponding to the six viewpoints. For each image, 2D object detection models for flowers are used to identify 2D bounding boxes which can be converted into the 3D space to extract flower point clouds. Pose estimation is performed by fitting three shapes (superellipsoids, paraboloids and planes) to the flower point clouds and compared with manually labeled ground truth.  Our method successfully finds approximately 80\% of flowers scanned using our customized FarmBot platform and has a mean flower pose error of 7.7 degrees, which is sufficient for robotic pollination and rivals previous results. All code will be made available at \href{https://github.com/harshmuriki/flowerPose.git}{https://github.com/harshmuriki/flowerPose.git}.

\end{abstract}

\section{Introduction}
\hspace{-\parindent}Urban farms \cite{usdaUrbanAg} provide healthy food to local communities and can serve as platforms for education and sustainability.  Unlike their rural counterparts, urban farms are usually small in scale and commercially available robotic systems such as the FarmBot \cite{FarmBot} have been developed to help automate basic cultivation tasks such as seeding, weeding, and watering.


The FarmBot (Fig.~1) is a scalable XYZ gantry robot that traverses linear rails, covering an arable area of up to 18 $\mbox{m}^2$ (0.0044 acre).  Such a platform makes it feasible to reach and potentially interact with each and every plant growing within its volumetric workspace.  We have used the FarmBot with a custom two degrees-of-freedom (DOF) camera end-effector to explore phenotyping for small-scale urban farms (though our methodology and results are applicable to other forms of controlled environment agriculture, such as indoor farming).  Specifically, this paper focuses on flower pose (position and orientation) estimation as a canonical phenotyping task, which is critical for robotic pollination.

\myfigure{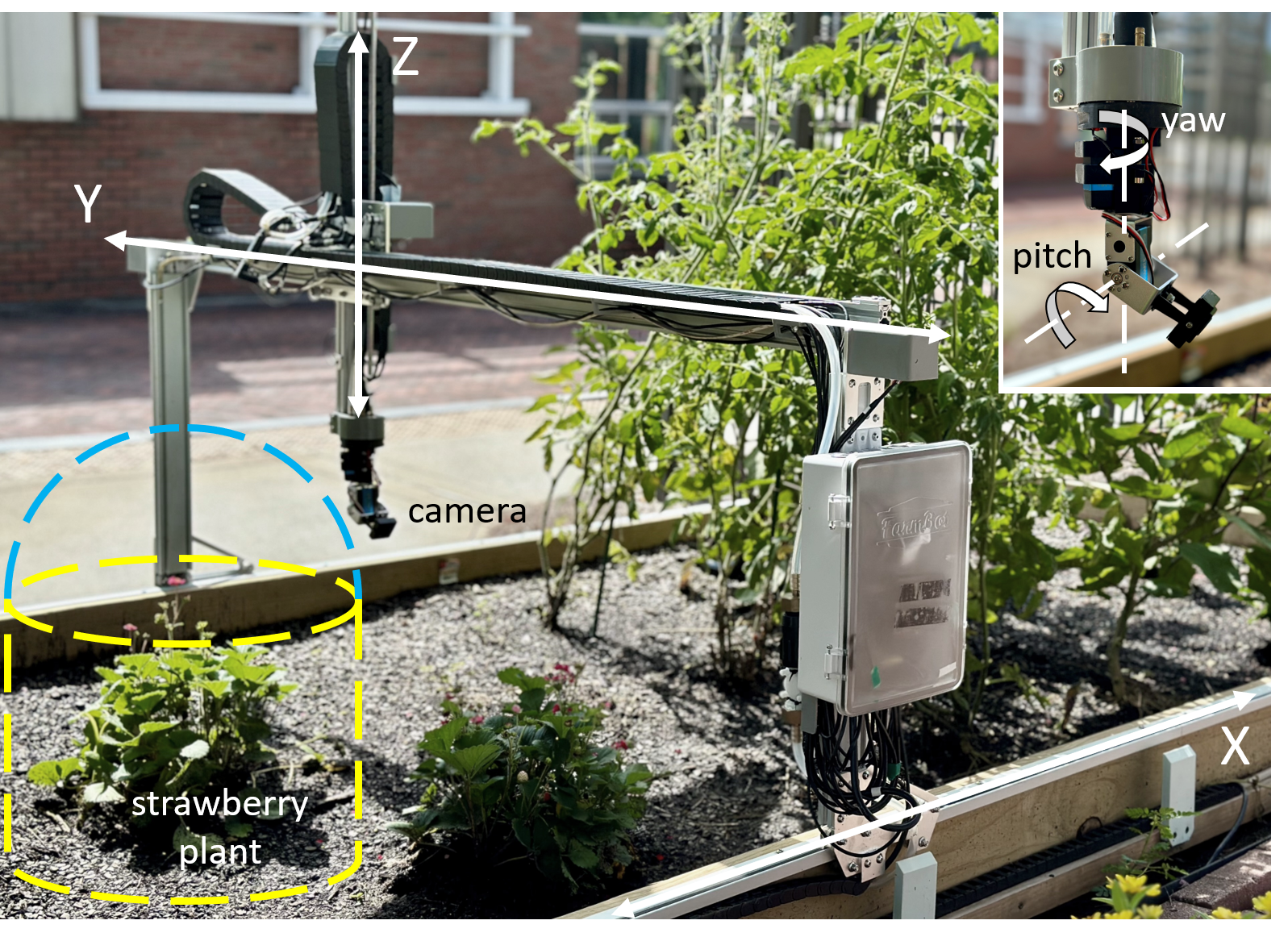}{FarmBot Genesis v1.7 tending a raised urban garden bed with strawberry plants in foreground.  Inset: custom 2-DOF camera end-effector.  The cylinder/dome plant scanning path for the camera is also schematically shown. \vspace{-20pt}}{fig:Farmbot}

Pollination is the transfer of pollen grains from the male stamen to the female pistil, either from one flower to another (cross-pollination) or from the same flower to itself (self-pollination).  The pistil is located in the center of the flower's petals, extending outwards roughly orthogonal to their surface, and surrounded by multiple stamen (each comprised of a filament supporting a pollen grain).  Knowledge of the pose of the flower petals therefore enables access to the stamen and pistil.
Robotic pollination can be used when natural pollinators such as bees are declining or unsuitable (as is the case for indoor farms).  Yuan et al. \cite{yuan2016} considered automated pollination for greenhouse tomatoes using a mobile robot arm platform and binocular vision for flower localization.
Smith et al. \cite{smith2024}, Strader et al. \cite{strader2019}, Yang et al. \cite{yang2019}, and Ohi et al.\cite{ohi2018} have implemented mobile robot arm platforms (including a multi-arm robot) for pollination of self-pollinating bramble plants in a greenhouse environment.  Flower detection is performed based on trained models (including YOLOv8 \cite{jocher2023}) using RGB images.  Three orientation classes (flower facing left, center and right) were learned in order to estimate flower pose for subsequent contact interaction between pollinator tool and flower center (containing the stamen and pistil).  Yang et al. \cite{yang2023} developed a similar approach, detecting forsythia flower pistils directly based on stereo images, while Ahmad et al. \cite{ahmad2024} worked with watermelons and used nine flower orientation classes (center and eight radially outward-pointing directions).  Finally, Hulens et al. \cite{hulens2022} developed an autonomous drone for flower pollination that trains its deep learning models on a mixed dataset of real, artificial and computer-rendered flowers at 10-degree angular increments about a vertical axis.


Prior research efforts in robotic pollination use discrete 2D images of flowers to infer orientation, with none providing a quantitative evaluation of angle estimation accuracy.  We leverage the capability of a customized FarmBot to reach arbitrary poses within a volumetric workspace enclosing a raised urban garden bed to autonomously generate 3D models of flowering strawberry plants.  A novel method is described to efficiently extract point clouds corresponding to the strawberry flowers for subsequent spatial pose estimation.  Our results are evaluated with respect to known ground truth pose values and will be used in our own work on robotic pollination \cite{kong2024} that relies on precise knowledge of flower pose.


\section{Method}
\subsection{Hardware Platform}

\hspace{-\parindent}We use a customized FarmBot Genesis v1.7 as our data acquisition platform. The XYZ gantry-type robot has been mounted onto a 5 ft $\times$ 10 ft raised garden bed (Fig.~1). A key feature of FarmBot is its universal tool mount (UTM), which acts as the robot's wrist that interfaces with custom tools via 12 pogo pins. The tools are held in place with neodymium magnets. The UTM can be positioned in the FarmBot's workspace using software commands issued through Farmbot's Python API \cite{farmbotAPI}.

The FarmBot's control system is composed of a Raspberry Pi \cite{raspberrypi} and a custom ATmega2560-based microcontroller (Farmduino). The Raspberry Pi runs FarmBot OS, which receives instructions from the FarmBot API via a cloud server. The Raspberry Pi in turn sends instructions to the Farmduino, which enables I/O pins to control different parts of the robot (twelve of these I/O pins connect with the UTM pogo pins).

A custom 2-DOF camera end-effector has been designed to be mounted onto the UTM, enabling yaw and pitch rotations
actuated via two high-torque 35 kg waterproof servo motors that are controlled using digital output pins D5 and D6 on the Farmduino.
The camera is an Arducam 4K 8MP IMX219 USB Autofocus Camera with 1280 $\times$ 720 video resolution.
To interface with the Arducam (via USB) and store recorded images/videos, we have added a separate Raspberry Pi 5 to the hardware set-up.


\subsection{Data Acquisition}
\hspace{-\parindent}Our approach to estimating flower pose is based on using autonomously generated 3D models of the supporting plant.  We have chosen flowering strawberry plants as the specimen of interest, with artificial white flowers used to augment our sample size.  Based on images taken of the sample plant from multiple camera poses, we use photogrammetry to stitch them together into a spatial model.

We have programmed the FarmBot's camera end-effector path to revolve around the plant in two parts: a cylinder to capture circumferential details topped by a hemisphere to capture top-down details (refer to the schematic in Fig. \ref{fig:Farmbot}). The camera, set at a fixed manual exposure, continuously records a video while traversing this scanning path. For the cylinder portion, the 2-DOF end-effector is actuated so that the camera points towards the cylinder's center-line, while for the half-dome it points towards the spherical center aligned with radial lines.
In order to take advantage of the auto-focus capability of the Arducam, the robot pauses at regular intervals along its scanning path to help ensure sharp and well-lit images of the flowering plant.

Once a video sequence of the flowering strawberry plant has been recorded (at 30 frames per second), images are extracted frame-by-frame.
As an example: for a plant with diameter 15 inches and height 12 inches, our scanning path takes approximately 20 minutes to execute.
This yields approximately 
36,000 image frames.
The images are separated into $N$ equally sized sequential bins, where $N$ is the number of images to be used for photogrammetry stitching.  Each bin corresponds to a segment of the scanning path, which ensures coverage of the entire plant.
For each bin, Laplacian filtering \cite{cv2Laplace} is performed on the central area of each image to obtain an average sharpness value, giving us a quality score to help choose the best image in the bin.


\subsection{3D Model Generation}
\hspace{-\parindent}We have used a commercially available off-the-shelf software tool called Polycam \cite{Polycam} to generate 3D photogrammetry models of the scanned flowering strawberry plants using the $N$ selected images.  Through empirical testing, we noticed that 200 images achieved dense point clouds with an average of 62,500 vertices, commensurate with the scale and detail of strawberry plants.

Additional images did not result in better quality point clouds. 
We used \textit{RAW} mode
and enabled the \textit{Sequential} option (since the images we captured were spatially ordered). To further increase the detail of the point cloud, we used Polycam's \textit{Remesh} feature, maximized the \textit{Polygon} count as well as \textit{Texture} resolution, and set the topology to \textit{Uniform}.
Fig. \ref{fig:polycam} shows an example 3D model output from Polycam, which includes the strawberry plant and the underlying soil surface.  We note that the flowers in our raised urban garden bed have white petals and yellow pistils.

\myfigureSmaller{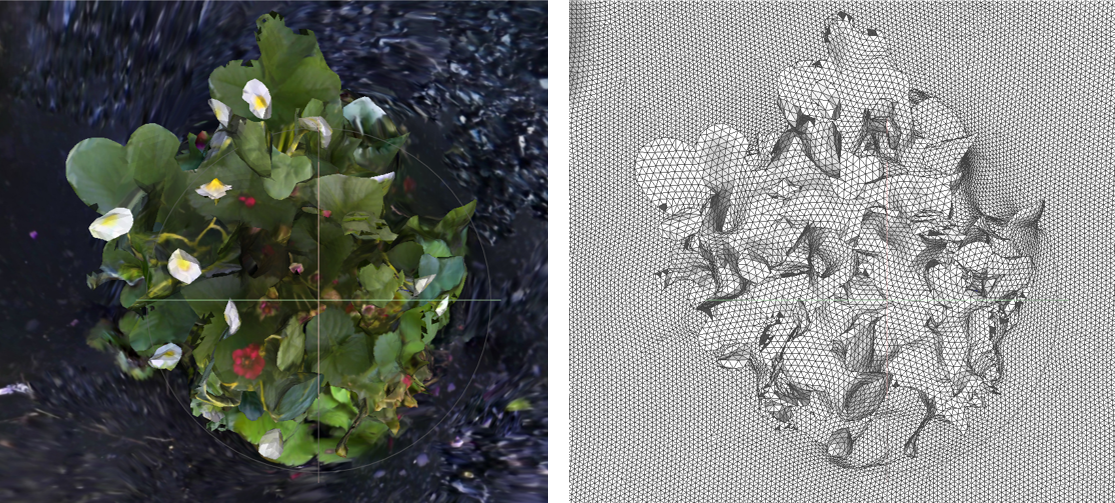}{Rendered (left) and wireframe (right) 3D mesh model of example flowering strawberry plant outputted from Polycam based on photogrammetry.\vspace{-15pt}}{fig:polycam}

\subsection{Translating Occupancy Grid Method}

\hspace{-\parindent}There has been extensive research on extracting 3D objects from point clouds, which can be broadly categorized into two main approaches: utilizing the points (or voxels) directly or projecting the points onto 2D planes. 3D deep learning models based on neural networks or transformer model architectures have been employed to work directly with points, like those discussed in Wu et al. \cite{wu2024} , Kolodiazhny et al. \cite{kolodiazhnyi2023} and Qi et al. \cite{qi2017pointnetplusplus}. However, one major limitation of this method is that their accuracy strongly depends on the dataset used during training. If a new class of objects is introduced, the model often requires re-training. This process is computationally intensive and can be costly, particularly when dealing with large datasets or complex models. Additionally, they require a very large amount of runtime because of the amount of data they work with.

\myfigureMedium{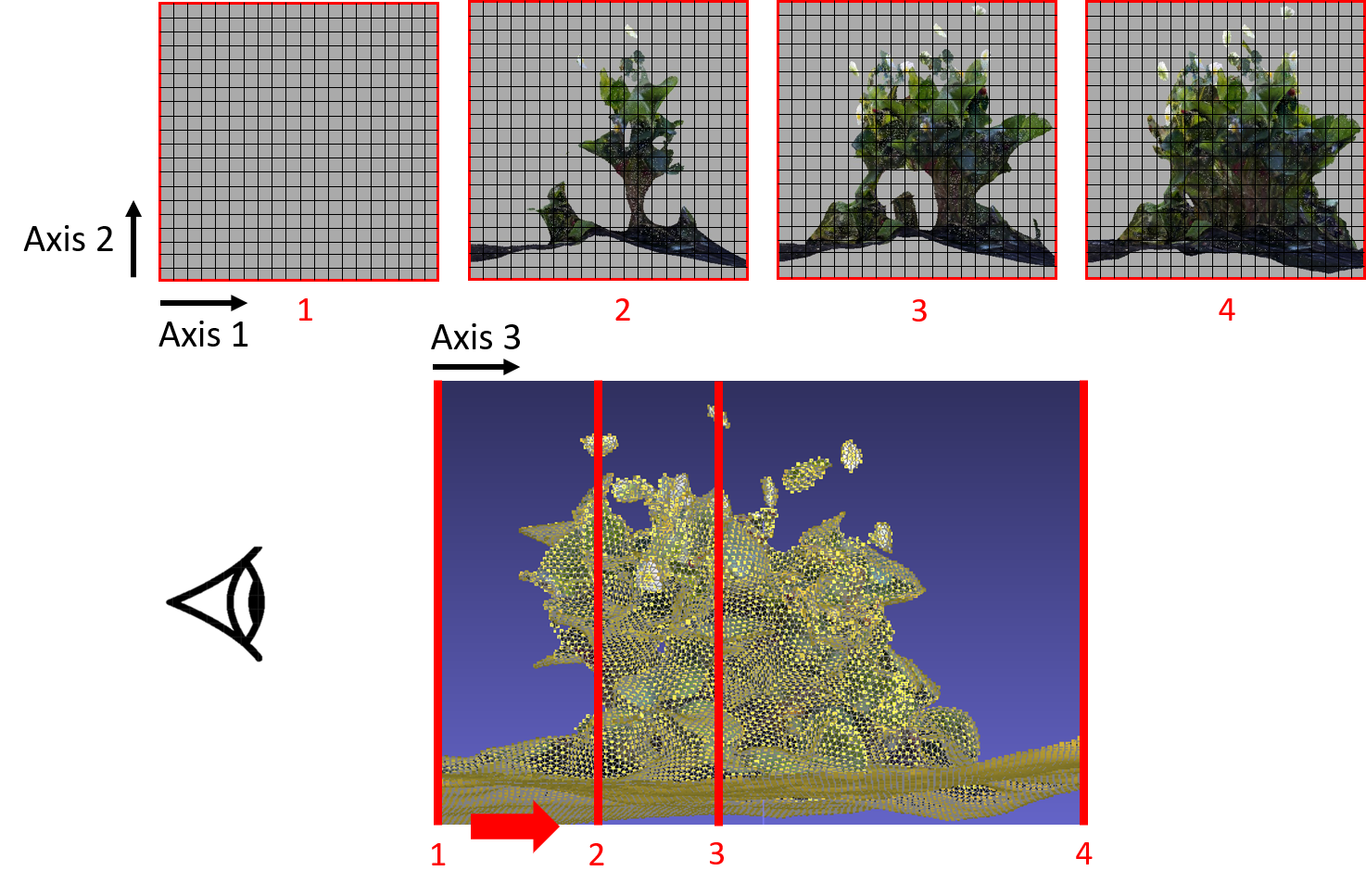}{Schematic of the translating occupancy grid that maps 3D points onto a 2D image, shown for one orthogonal viewpoint direction. \vspace{-20pt}}{fig:TranslatingOccupancyGridMethod}

\SetKwFunction{GetThreeDFromTwoD}{get3DPointcloudfrom2dImage}
\SetKwFunction{GetTwoDFromThreeD}{get2dImagefrom3dPointcloud}
\SetKwFunction{TranslatingOccupancyGridMethod}{TranslatingOccupancyGridMethod}

\setlength{\algomargin}{0em} 
\begin{algorithm}
\small
\caption{Pseudocode for Translating Occupancy Grid Method}
\SetKwFunction{FuncName}{TranslatingOccupancyGridMethod}
\label{algo:TranslatingOccupancyGridMethod}
\FuncName{}

\KwData{Pointcloud}
\KwResult{Array of 3D coordinates of detected strawberry flowers}

\texttt{orthogonalSides} = $initialization(\texttt{Pointcloud})$

\texttt{Pointcloud} = $threshold$\texttt{(Pointcloud)} \texttt{\{Remove ground points using bounding boxes\}}

\For{each \texttt{side} in \texttt{orthogonalSides}}{

    \texttt{2Dimage, 2Dgrid, 2Ddata} = 
    \hspace{1em} \GetTwoDFromThreeD{\texttt{Pointcloud, side}}
    
    \texttt{boundingBoxes} = $objectDetection$(\texttt{2Dimage})

    \texttt{allData} = \texttt{[]}
    
    \For{each \texttt{boundingBox} in \texttt{boundingBoxes}}{
    
        \texttt{3Ddata} = \GetThreeDFromTwoD{
        \hspace{5em} \texttt{boundingBox, 2Dgrid, 2Ddata}}
        
        \texttt{allData} += \texttt{3Ddata}

    }

    \texttt{newPcd} = \texttt{New pointcloud with $allData$ points}

}
\texttt{newPcd = apply statistical and radius}

\texttt{\hspace{5em} outlier removal on $\texttt{newPcd}$}

\texttt{boundingBoxes, segments} =  $DBSCAN(\texttt{newPcd})$

\texttt{\textbf{return}} \texttt{segments}
\end{algorithm}






    




Utilizing a projection-based system where point clouds are projected directly onto 2D planes addresses some of these limitations. For example, Boulch et al. \cite{boulch2018} employed manually captured 2D snapshots of different views of the point cloud to segment objects. In contrast, Lahoud et al. \cite{lahoud2017} use a single RGB-D image to estimate the location of an object. Another approach by Yang et al. \cite{yang2020} uses a mini T-net model to extract features from 3D point clouds and project them onto 2D planes. Lyu et al. \cite{lyu2020} uses Delaunary triangulation to convert 3D point clouds to 2D images.
However, these approaches perform best when the detected object constitutes a large fraction of the point cloud with relatively little external noise. Furthermore, some of these processes are either very laborious to execute \cite{boulch2018} or have low accuracy. Moreover, the downstream task of pose estimation necessitates 3D positional information, which cannot be derived solely from 2D images and detections; hence, generating and processing a point cloud is essential.

\setlength{\intextsep}{5pt}
\begin{algorithm}
\small
\caption{Pseudocode for 3D to 2D Conversion}
\label{algo:get2dImagefrom3dPointcloud}
\SetKwFunction{FuncName}{get2dImagefrom3dPointcloud}
\FuncName{}

\KwData{Pointcloud, side}
\KwResult{Array of image, grid, and data}

\texttt{resolution} = \texttt{10} 

\texttt{width, height} = \texttt{2D\_Image.shape}


\texttt{N = \texttt{width} + 2 $\times$ \texttt{resolution}}

\texttt{M = \texttt{height} + 2 $\times$ \texttt{resolution}}

\texttt{checkGrid} = \texttt{Array of zeros of shape (N,M)}

 \texttt{grid} = \texttt{Array of zeros of shape (N,M)}

 \texttt{image} = \texttt{A blank grayscale RGB image}

 \texttt{data} = \texttt{\{empty dictionary\}}

\texttt{axis1, axis2} = \texttt{axes}



\texttt{points} = \texttt{sort(points, axis3)}

\texttt{colors} = \texttt{sort(colors, axis3)}



\For{each \texttt{(idx,coords)} in \texttt{points}}{
    
    \texttt{coords\_norm} = \texttt{normalize coords to [-1,1]}
    \hspace{3.5em} \texttt{along (axis1, axis2)} 
    
    
    \texttt{index1} = \texttt{(coords\_norm.axis1 * width +}
    \hspace{3.5em}\texttt{(width / 2)) + resolution}
    
    \texttt{index2} = \texttt{(coords\_norm.axis2 * height +}
    \hspace{3.5em}\texttt{(height / 2)) + resolution}

    \If{not \texttt{checkGrid[index1,index2]}}{
        \texttt{checkGrid[index1,index2]} = \texttt{1}
        
        \texttt{data[idx]} = [\texttt{colors[idx]}, \texttt{coords}]
        
        \texttt{grid[index1,index2]} = \texttt{idx}
        
        \texttt{image[index1,index2]} = \texttt{colors[idx]}
        
        }
    }
    
\texttt{\textbf{return}} \texttt{image,  grid, data}

\end{algorithm}

We have developed a novel method for computing 3D bounding boxes around flowers in the point cloud of the strawberry plant generated in the previous section, summarized in Algorithm \ref{algo:TranslatingOccupancyGridMethod}. Six orthogonal snapshots were captured, each representing different perspectives of the plant, using open-source tools such as Open3d \cite{zhou2018}, OpenCV \cite{opencv} and Numpy \cite{numpy}. This allowed us to capture all the visible flowers from every side of the plant. Using these six orthogonal sides, we had six directions along three orthogonal axes: $X$, $-X$, $Y$, $-Y$, $Z$, and $-Z$. We then used a 2D occupancy grid as seen in Fig. \ref{fig:TranslatingOccupancyGridMethod} that moves along each of these 6 directions.
Each square element within the grid was assigned based on the first 3D point it encountered, as illustrated in the progression from location 1 to 4 in Fig. \ref{fig:TranslatingOccupancyGridMethod}.  The occupancy grid from location 4 (upon exiting the point cloud) is the resulting 2D image. We note that each detected element in the occupancy grid retains an associated ``depth'' along the viewpoint direction.


This process is detailed in Algorithm~\ref{algo:get2dImagefrom3dPointcloud}.
It is used to generate six 2D color images, illustrated in Fig. \ref{fig:6_bbox} as the sides of a cube (we used square occupancy grids).
Each grid element is essentially a pixel.
An important detail to note is the resolution parameter in Algorithm \ref{algo:get2dImagefrom3dPointcloud}. This parameter defines the radius, in pixels, around each detected (occupied) pixel. All pixels within this radius are colored the same color as the detected pixel. The resolution can be adjusted to modify the 2D image's granularity. Our empirical evidence demonstrates that capturing high-resolution snapshots (e.g., 7000 $\times$ 7000 pixels) of point clouds containing numerous objects, such as plants, produces highly detailed images with minimal pixelation.

\begin{algorithm}
\small
\caption{Pseudocode for 2D to 3D Conversion}
\SetKwFunction{FuncName}{get3DPointcloudfrom2dImage}
\label{algo:get3DPointcloudfrom2dImage}
\FuncName{}

\KwData{boundingBox, grid, data}
\KwResult{Array of 3D points and colors of only flowers}

\texttt{x\_min, x\_max, y\_min, y\_max} = \texttt{boundingBox}

 \texttt{colors, points} = \texttt{[]}

\For{each \texttt{x} in \texttt{range(x\_min,x\_max,1)}}{

    \For{each \texttt{y} in \texttt{range(y\_min,y\_max,1)}}{
    
        \texttt{(colors, points)} = \texttt{(colors, points) $\cup$}
        \hspace{4.1em} \texttt{(colors, 3D coordinate at (x,y))}

        

        }

    }
    
\texttt{\textbf{return} \texttt{(points, colors)}} 

\end{algorithm}

Multiple object detection algorithms such as pre-trained CNNs, color-based thresholding, etc. can be employed due to the modular nature of the process. We decided to leverage pre-trained object detection models like YOLOv10 \cite{yolov10} and Roboflow 3.0 Object Detection \cite{Roboflow_Model}, fine-tuning them to detect white strawberry flowers.
As Algorithm~\ref{algo:TranslatingOccupancyGridMethod} illustrates, each image generates multiple 2D bounding boxes since there are numerous detected flowers. Since Algorithm~\ref{algo:get2dImagefrom3dPointcloud} mapped each pixel to its corresponding 3D coordinate, we can use these 2D bounding boxes to extract the 3D points associated with the detected flowers, as described in Algorithm~\ref{algo:get3DPointcloudfrom2dImage}. We then merge all of these 3D points into a unified 3D point cloud and apply density-based spatial clustering of applications with noise (DBSCAN) \cite{dbscan} clustering based on spatial location to segment and filter the flowers from the noise. Through domain knowledge and iterative experimentation, we set the parameters \textsl{eps} and \textsl{min\_points} to 0.01 and 20 respectively.
Smaller \textsl{eps} leads to fragmented clusters while higher \textsl{min\_points} reduces sensitivity to noise and excludes some of the smaller clusters.
This resulted in the creation of 3D bounding cuboids around the strawberry flowers, as shown in Fig.~\ref{fig:process} (bottom left). This method is computationally faster and more flexible than aforementioned work.

\subsection{Flower Pose Estimation}
\subsubsection{Post-Processing}
Using the bounding cuboids outputted from the Translating Occupancy Grid Method, we obtain the point clouds for each of the flowers found in the strawberry plant.
For each flower, we seek to extract the points in the point cloud corresponding to the petals and the pistil separately, since flower poses are intuited from how the corolla (the petals collectively) opens up and from the position of the pistil. Since the strawberry flowers used
have distinct white petals and yellow pistils, we accomplish this by converting the point colors into the hue, saturation, value (HSV) color space and applying separate threshold filters for the petals and the pistil. This method successfully segments out the petals. However, due to the yellowish-green color of the flower stems,
the threshold filter for the pistil includes some spurious points from the base of the flower.

To address this issue,
we perform DBSCAN clustering on the result of the pistil threshold filter, with each of the clusters generated being potential candidates for the pistil. DBSCAN groups together points that are close to each other based on two parameters: a distance metric and the minimum number of points contained in each cluster. These parameters have been set keeping real world flowers in mind. Based on the assumption that the flower pistil should be located in the middle of the petals, we select the cluster whose centroid has the shortest Euclidean distance to the centroid of the petal point cloud obtained in the previous step. Fig.~\ref{fig:process} shows the entire process of extracting the petals and pistil for each of the flowers.

\myfigureMedium{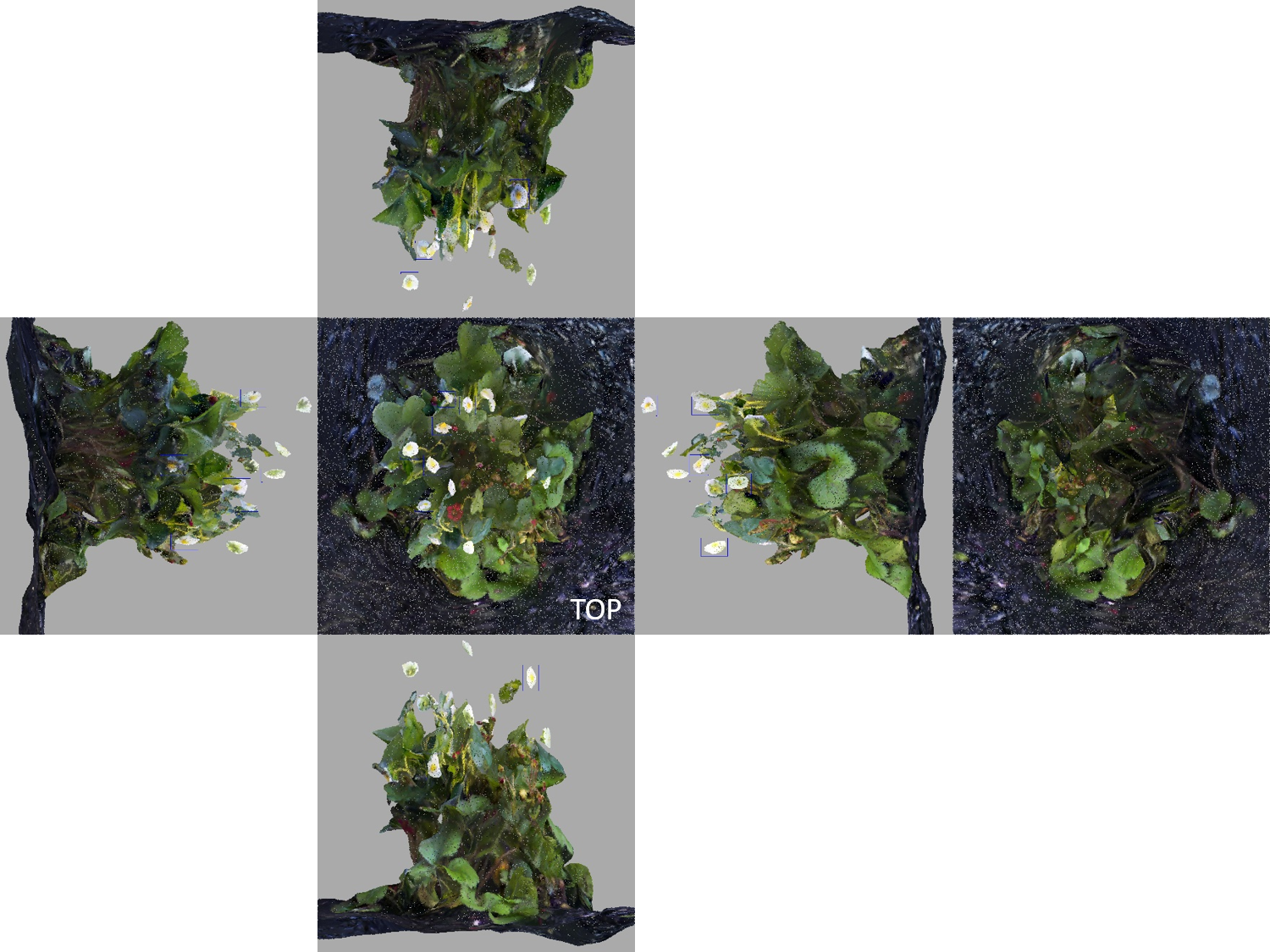}{Images resulting from the translating occupancy grid method applied along six orthogonal views.\vspace{-5pt}}{fig:6_bbox}

\myfigure{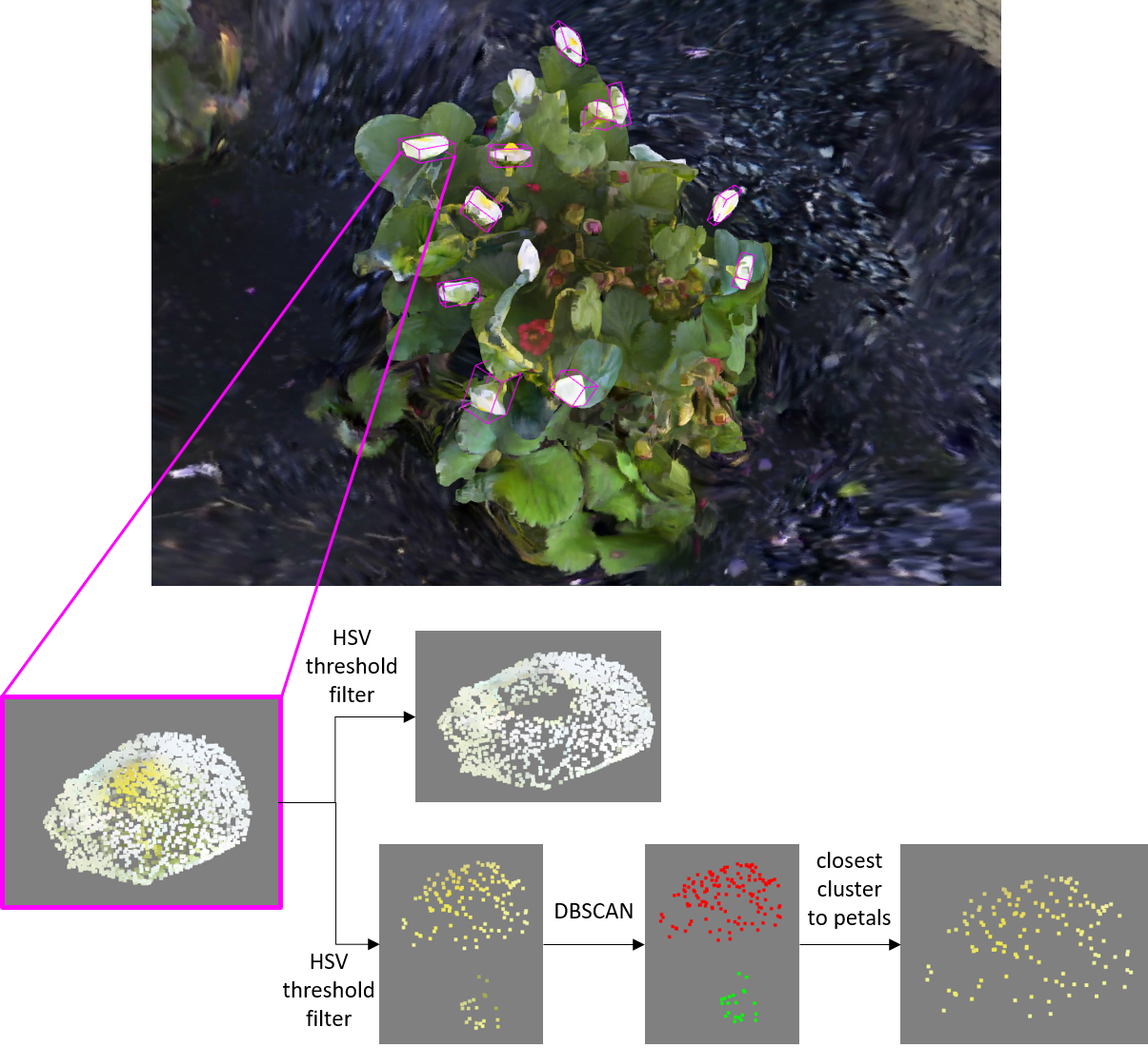}{Illustration of proceeding from bounding cuboids from translating occupancy grid method to flower point cloud (minus pistils, etc.).\vspace{-17pt}}{fig:process}

\subsubsection{Shape Fitting}
The extracted petal points will be used to determine flower pose.
After centering the petal point cloud at its centroid (origin), we proceed to fit quadric surfaces.
Three candidate surfaces were investigated: superellipsoid (Eq.~\ref{eq:superellipsoid}),
paraboloid ($z = (x/a)^2+(y/b)^2$)
and plane ($ax + by + cz =d$).

\begin{equation}
\left[\left(\frac{x}{a}\right)^\frac{2}{\epsilon_2} + \left(\frac{y}{b}\right)^\frac{2}{\epsilon_2}\right]^\frac{\epsilon_2}{\epsilon_1} + \left(\frac{z}{c}\right)^\frac{2}{\epsilon_1} = 1
\label{eq:superellipsoid}
\end{equation}

The motivation for fitting a \textit{superellipsoid}
comes from the observation that the petal point cloud often traces out part of a rounded and closed surface which is similar to that of an ellipsoid. Superellipsoids are the generalization of an ellipsoid that can be additionally deformed
using two introduced parameters, adding greater shape-fitting flexibility (e.g., see \cite{bellpepper, bellpepper2, dough}).
A superellipsoid is fit onto the petal point cloud by performing constrained nonlinear least-squares optimization
on parameters $a$, $b$, $c$, $\epsilon_1,$ and $\epsilon_2$ as well as on Euler angles $\phi$, $\theta$ and $\psi$ that define the spatial orientation of the superellipsoid. The optimization is performed by the trust region reflective algorithm \cite{trf} which solves a system of equations motivated by a first-order optimality condition. 
The bounds on the parameters are: $0 \leq a, b, c \leq 0.1$ and
$0.9 \leq \epsilon_1, \epsilon_2 \leq 1.1$.
Parameters
$a$, $b$ and $c$ are bounded to be smaller than 0.1 m to reflect the size of real-life flowers, while $\epsilon_1$ and $\epsilon_2$ are bounded so that the shape of the superellipsoid is similar
to one that empirically yields the best result compared to having no bounds or to directly fitting an ellipsoid.


The next step is to determine which of the six directions in the coordinate frame of the superellipsoid (two for each axis) is the pose estimate for the flower. We achieve this by finding the minimum of $a$, $b$ and $c$ (which, similar to an ellipsoid, describe how much the superellipsoid is stretched in each axis) to find the shortest axis spanned by the superellipsoid. This axis would be parallel to the pose estimate.
Finally, to determine the correct direction along that axis, we use the pistil point cloud extracted in the previous step. Since the pistil should always be on the side of the flower where the corolla opens up (opposite the stem), we define a vector originating from the centroid of the entire flower point cloud to the centroid of the pistil point cloud. The direction whose dot product with the aforementioned vector is positive is chosen as the estimated flower pose.

The motivation for fitting a \textit{paraboloid}
is the observation that the petals also had a tendency to curve upward along the sides of the flower. The advantage of using a paraboloid over a superellipsoid is the fact that it is directional, making it a trivial task to determine the pose estimate once the fitting is performed. Similar to the superellipsoid, we fit a paraboloid onto the petal point cloud by performing unconstrained nonlinear least-squares optimization on parameters $a$ and $b$, as well as Euler angles $\phi$, $\theta$ and $\psi$. The Levenberg-Marquardt nonlinear least-squares algorithm \cite{gavin2013} is used for optimization. 
Since the paraboloid is directional along the z-axis of its local coordinate frame, the pose estimate is simply the direction of the positive z-axis transformed into the global coordinate frame. This offers an advantage over the superellipsoid where determining which of the six directions to use as the pose estimate is a non-trivial task.

Finally, we experimented with fitting a \textit{plane} because we noticed that the superellipsoid and paraboloid can occasionally be skewed by outliers in the point cloud. A plane is a simpler mathematical model that would be less prone to over-fitting.
We perform the fitting by applying principal component analysis to the petal point cloud and selecting the eigenvector corresponding to the smallest eigenvalue. This eigenvector represents the direction with the smallest variation in the point cloud data and thus it would be parallel to the normal of the best-fit plane.
Then, just as for the superellipsoid estimate, the pistil of the flower is used to choose the correct direction between the two possible normals (180 degrees apart) to the plane.

For illustration, Fig.~\ref{fig:shapes} shows an example result of fitting all three quadrics onto a petal point cloud. Colored arrows represent the respective three pose estimates. The black arrow denotes ground truth pose.

\myfigureSmaller{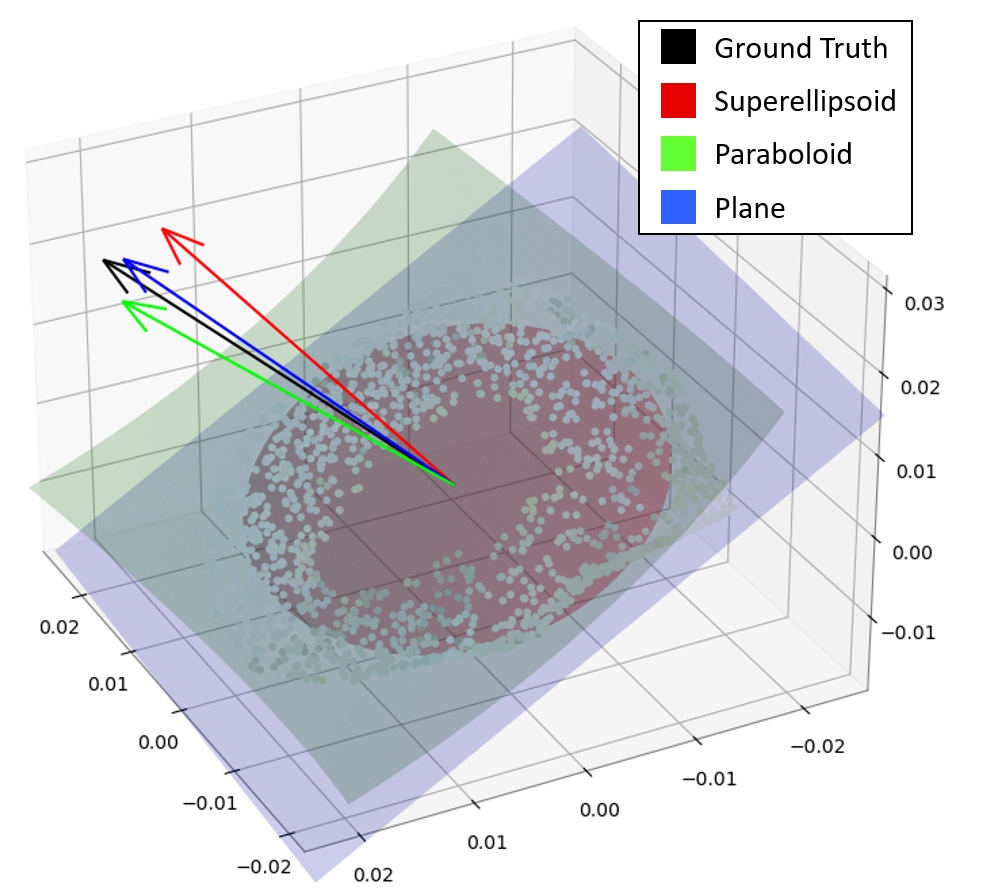}{Example of ground truth pose as well as superellipsoid, paraboloid and plane pose estimates on a flower.\vspace{-20pt}}{fig:shapes}

\vspace{-3pt}
\section{Results}
\hspace{-\parindent}We used the customized FarmBot platform to autonomously acquire seven scans of flowering strawberry plants. 
This section outlines the process of deriving ground truth and estimated flower poses from scans from our methodology.


{\subsection{Ground Truth Specification of Flower Pose}}
\hspace{-\parindent}We annotate by hand the ``ground truth'' pose of each flower in the
scans using AWS SageMaker Ground Truth \cite{aws} software. This is done by adding oriented bounding boxes with arrows on discerned flowers,
as seen in Fig.~\ref{fig:aws}. Cases which cannot be labeled include concealed flowers or ones with heavily distorted corollas, which resulted from lower quality point clouds due to the Arducam's low resolution (read section III-B for improvements). We estimate 5° of human error in ground truth pose values. The number of flowers labeled with the ground truth pose for each plant scan is shown in the second column of TABLE \ref{tab:resultsFarmBot}.

\myfigureSmaller{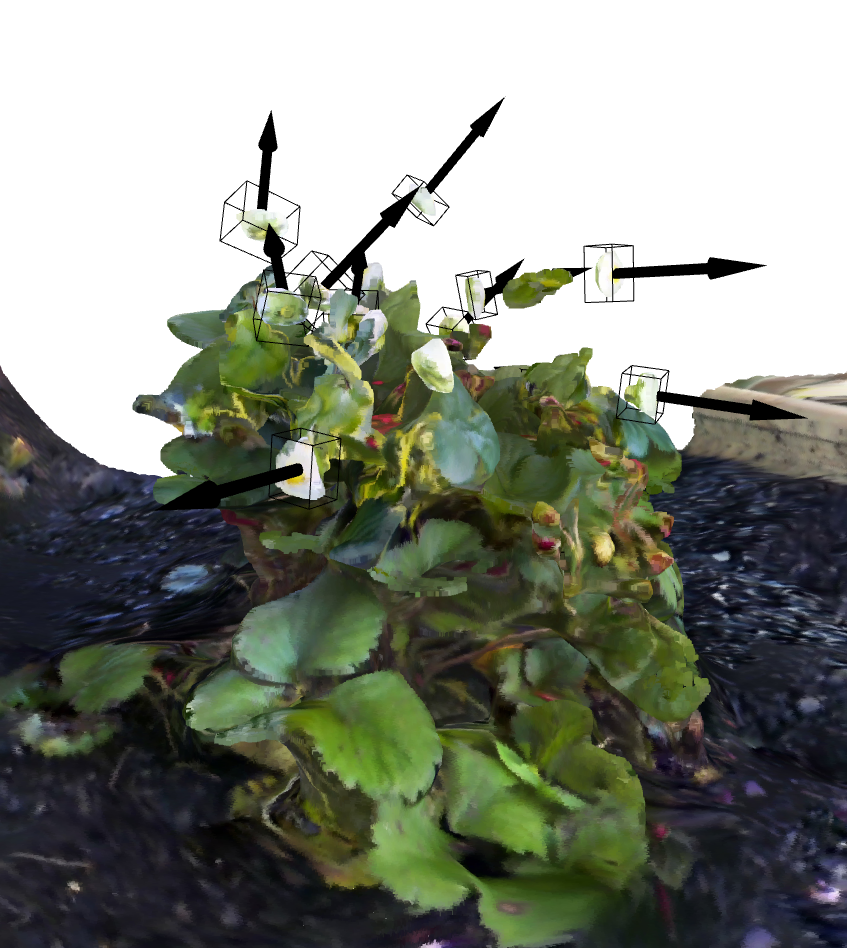}{Example of ground truth flower pose labeling using AWS SageMaker Ground Truth.\vspace{-17pt}}{fig:aws}

\vspace{-3pt}

\subsection{Flower Pose Estimation Results}

\hspace{-\parindent}We estimate the pose for a flower using the three quadric surfaces described in section II.E. only if it has been detected using the Translating Occupancy Grid Method and has been assigned a ground truth pose value. The results appear in TABLE~\ref{tab:resultsFarmBot}. Overall, 80.3\% of ground truth flowers were detected using our method.
An edge case occurred with Plant ID 6, where only 50\% of the flowers were detected. This is because, during the object detection phase, the model failed to detect some of the flowers due to merging with leaves or being occluded.

We find that the plane performs the best with a mean error of 7.7 degrees, which is a
satisfactory
result given that the ground truth itself could be off by 5 degrees due to human error. There are two main sources of error, both of which are caused by poor point clouds. Firstly, the petals can curve downward and fuse with the stem, resulting in a mismatch with the mathematical model of a flat plane. Secondly, the white flower petals can fuse with the pistil and take on a yellow color, causing it to be filtered out during petal extraction.

The superellipsoid (the next best method)
suffers from the same problems as the plane. This shape also occasionally does not fit properly to the petal point cloud, resulting in significant pose errors. We hypothesize that this is because the algorithm gets stuck in a local minima since the objective function in the least-squares optimization is non-convex.

The paraboloid has the worst performance because it is based on the assumption that the flower petals curve upward. However (as mentioned previously) a number of flowers have downward curving petals, resulting in a pose estimation that can be 180° off from the ground truth pose. The error in these cases is very large and explains the large disparity between the mean and the median.

There are a few cases where a flower is detected by the Translating Occupancy Grid Method but has not been assigned a ground truth pose, labeled as `No. Extra' in TABLE \ref{tab:resultsFarmBot}. This is because these particular poses are difficult to manually determine due to a deformation in the point cloud or occlusion from foliage, but is detected by the algorithm since there remains a resemblance to a strawberry flower. The cases where a detected ``flower" is actually a leaf or part of another flower that has already been detected are labeled as `FP' (false positives) in TABLE~\ref{tab:resultsFarmBot}.


\begin{table}
    \centering
    \begin{tabular}{p{0.65cm}|p{0.33cm}|p{1.1cm}|p{0.61cm}|p{0.25cm}|p{0.8cm}|p{0.57cm}|p{0.55cm}}
        \hline
        Plant ID & GT & Flowers Found (\%) & No. Extra & FP & Super-ellipsoid error & Para-boloid error & Plane error \\ \hline \hline
        1 & 7 & 6 (85.7) & 1 & 0 & 14.7° & 34.7° & 5.4° \\ \hline
        2 & 10 & 10 (100) & 2 & 1 & 12.1° & 60.5° & 9.5° \\ \hline
        3 & 11 & 9 (81.8) & 3 & 2 & 26.7° & 10.2° & 8.0° \\ \hline
        4 & 7 & 6 (85.7) & 5 & 0 &  19.5° & 30.2° & 9.6° \\ \hline
        5 & 9 & 7 (77.8)& 2 & 0 &  42.0° & 120.3° & 7.4° \\ \hline
        6 & 10 & 5 (50) & 4 & 1 &  8.0° & 172.0° & 7.5° \\ \hline
        7 & 7 & 6 (85.7)& 6 & 0 & 7.5° & 7.9° & 4.6° \\ \hline \hline
        \textbf{Total} & 61 & 49 (80.3) & 23 & 4 & - & - & - \\ \hline
        \textbf{Mean} & - & - & - & - &  19.3° & 57.9° & 7.7° \\ \hline
        \textbf{Med} & - & - & - & - & 8.9° & 11.8° & 5.9° \\ \hline
        \textbf{Std Dev} & - & - & - & - & 33.6° & 73.3° & 5.8° \\ \hline
    \end{tabular}
    \caption{Flower pose estimation errors for each plant scan captured using the FarmBot platform.  GT: ground truth.  FP: false positives.\vspace{-20pt}}
    \label{tab:resultsFarmBot}
\end{table}

Cameras aren't perfect. Since many of our errors arise from the relatively poor quality of scans obtained using the Arducam, we manually captured an eighth scan using a smartphone with a higher resolution rear camera that produced a 3D model with 125,000 vertices. This is about twice the resolution of the autonomous FarmBot scans. The improved results for this scan are presented in TABLE~\ref{tab:resultsIphone}.  Note the significantly higher number of point cloud flowers that can be labeled with a ground truth pose, due to better 3D mesh models that facilitate manual pose labeling. 

\begin{table}[h]
    \centering
    \begin{tabular}{p{0.65cm}|p{0.33cm}|p{1.1cm}|p{0.61cm}|p{0.25cm}|p{0.8cm}|p{0.57cm}|p{0.55cm}}
        \hline
        Plant ID & GT & Flowers Found (\%) & No. Extra & FP & Super-ellipsoid error & Para-boloid error & Plane error \\ \hline \hline
        8 & 19& 15 (78.9) & 0& 2 & 17.5& 41.6& 6.4\\ \hline \hline
    \end{tabular}
    \caption{Mean flower pose estimation errors for the plant scan captured using a smartphone.}
    \label{tab:resultsIphone}
    \vspace{-3pt}
\end{table}

\section{Conclusions and Future Work}
\hspace{-\parindent}Through automated data acquisition using a customized FarmBot, point cloud 
reconstruction by Polycam, and application of our novel method of extracting 
flowers from point clouds, we have successfully developed a pipeline to 
estimate flower pose that finds approximately 80\% of flowers scanned
using the gantry-style robotic platform.  The mean flower pose error for pollination is 7.7 degrees, which is sufficient for robotic pollination and rivals previous results such as Ci et al. \cite{ci}, Sun et al. \cite{sun} and Luo et al. \cite{luo}.

Using a gantry robot places limits on plant height and farm area, making our system most suitable to smaller urban farms. The parameters for the Translating Occupancy Grid Method are also specific to strawberry flowers and would have to be tuned for other kinds of flowers.

As for future work, we should be able to increase the percentage of flowers found by using semantic segmentation instead of object detection. This is because object detection yields a 2D rectangular bounding box, which includes unwanted noise artifacts whereas semantic segmentation does not. Furthermore, using an RGB-D instead of an RGB camera would require fewer images to generate a good 3D model of the plant, speeding up the data acquisition step.

\newpage
\small
\bibliographystyle{ieeetr} 
\bibliography{references}

\begin{thebibliography}{10}

\bibitem{usdaUrbanAg}
{United States Department of Agriculture}, ``Urban agriculture.'' \url{http://www.nal.usda.gov/farms-and-agricultural-production-systems/urban-agriculture}.

\bibitem{FarmBot}
{FarmBot}, ``Farmbot.'' \url{http://farm.bot}.

\bibitem{yuan2016}
T.~Yuan, S.~Zhang, X.~Sheng, D.~Wang, Y.~Gong, and W.~Li, ``An autonomous pollination robot for hormone treatment of tomato flower in greenhouse,'' in {\em 2016 3rd International Conference on Systems and Informatics (ICSAI)}, pp.~108--113, IEEE, 2016.

\bibitem{smith2024}
T.~Smith, M.~Rijal, C.~Tatsch, R.~M. Butts, J.~Beard, R.~T. Cook, A.~Chu, J.~Gross, and Y.~Gu, ``Design of stickbug: a six-armed precision pollination robot,'' {\em arXiv preprint}, 2024.

\bibitem{strader2019}
J.~Strader, J.~Nguyen, C.~Tatsch, Y.~Du, K.~Lassak, B.~Buzzo, R.~Watson, H.~Cerbone, N.~Ohi, C.~Yang, and Y.~Gu, ``Flower interaction subsystem for a precision pollination robot,'' in {\em 2019 IEEE/RSJ International Conference on Intelligent Robots and Systems (IROS)}, pp.~5534--5541, IEEE, 2019.

\bibitem{yang2019}
C.~Yang, R.~M. Watson, J.~N. Gross, and Y.~Gu, ``Localization algorithm design and evaluation for an autonomous pollination robot,'' in {\em International Meeting of The Satellite Division of the Institute of Navigation}, pp.~2702--2710, 2019.

\bibitem{ohi2018}
N.~Ohi, K.~Lassak, R.~Watson, J.~Strader, Y.~Du, C.~Yang, G.~Hedrick, J.~Nguyen, S.~Harper, D.~Reynolds, and C.~Kilic, ``Design of an autonomous precision pollination robot,'' in {\em 2018 IEEE/RSJ International Conference on Intelligent Robots and Systems (IROS)}, pp.~7711--7718, IEEE, 2018.

\bibitem{jocher2023}
G.~Jocher, A.~Chaurasia, and J.~Qiu, ``Ultralytics yolo.'' \url{https://github.com/ultralytics/ultralytics}, 2023.

\bibitem{yang2023}
M.~Yang, H.~Lyu, Y.~Zhao, Y.~Sun, H.~Pan, Q.~Sun, J.~Chen, B.~Qiang, and H.~Yang, ``Delivery of pollen to forsythia flower pistils autonomously and precisely using a robot arm,'' {\em Computers and Electronics in Agriculture}, vol.~214, p.~108274, 2023.

\bibitem{ahmad2024}
K.~Ahmad, J.~E. Park, T.~Ilyas, J.~H. Lee, J.~H. Lee, S.~Kim, and H.~Kim, ``Accurate and robust pollinations for watermelons using intelligence guided visual servoing,'' {\em Computers and Electronics in Agriculture}, vol.~219, p.~108753, 2024.

\bibitem{hulens2022}
D.~Hulens, W.~Van~Ranst, Y.~Cao, and T.~Goedem{\'e}, ``Autonomous visual navigation for a flower pollination drone,'' {\em Machines}, vol.~10, no.~5, p.~364, 2022.

\bibitem{kong2024}
C.~Kong, A.~Qiu, I.~Wibowo, M.~Ren, A.~Dhori, K.-S. Ling, A.-P. Hu, and S.~Kousik, ``Towards closing the loop in robotic pollination for indoor farming via autonomous microscopic inspection,'' 2024.
\newblock In preparation.

\bibitem{farmbotAPI}
{FarmBot Developer API}, ``Farmbot developer api.'' \url{http://developer.farm.bot/v15/docs/web-app/rest-api.html}.

\bibitem{raspberrypi}
{Raspberry Pi Foundation}, ``Raspberry pi.'' \url{https://www.raspberrypi.com/}, 2024.

\bibitem{cv2Laplace}
{OpenCV}, ``Laplace operator.'' \url{http://docs.opencv.org/4.x/d5/db5/tutorial_laplace_operator.html}.

\bibitem{Polycam}
{Polycam}, ``Polycam.'' \url{http://poly.cam}.

\bibitem{wu2024}
X.~Wu, L.~Jiang, P.~Wang, Z.~Liu, X.~Liu, Y.~Qiao, W.~Ouyang, T.~He, and H.~Zhao, ``Point transformer v3: Simpler, faster, stronger,'' {\em arXiv preprint}, 2024.

\bibitem{kolodiazhnyi2023}
M.~Kolodiazhnyi, A.~Vorontsova, A.~Konushin, and D.~Rukhovich, ``Oneformer3d: One transformer for unified point cloud segmentation,'' {\em arXiv preprint}, 2023.

\bibitem{qi2017pointnetplusplus}
C.~R. Qi, L.~Yi, H.~Su, and L.~J. Guibas, ``Pointnet++: Deep hierarchical feature learning on point sets in a metric space,'' {\em arXiv preprint}, 2017.

\bibitem{boulch2018}
A.~Boulch, J.~Guerry, B.~Le~Saux, and N.~Audebert, ``Snapnet: 3d point cloud semantic labeling with 2d deep segmentation networks,'' {\em Computers \& Graphics}, vol.~71, pp.~189--198, 2018.

\bibitem{lahoud2017}
J.~Lahoud and B.~Ghanem, ``2d-driven 3d object detection in rgb-d images,'' in {\em Proceedings of the IEEE/CVF International Conference on Computer Vision (ICCV)}, King Abdullah University of Science and Technology (KAUST), Thuwal, Saudi Arabia, 2017.

\bibitem{yang2020}
J.~Yang, C.~Lee, P.~Ahn, H.~Lee, E.~Yi, and J.~Kim, ``Pbp-net: Point projection and back-projection network for 3d point cloud segmentation,'' in {\em 2020 IEEE/RSJ International Conference on Intelligent Robots and Systems (IROS)}, pp.~8469--8475, IEEE, 2020.

\bibitem{lyu2020}
Y.~Lyu, X.~Huang, and Z.~Zhang, ``Learning to segment 3d point clouds in 2d image space,'' in {\em Proceedings of the IEEE/CVF Conference on Computer Vision and Pattern Recognition (CVPR)}, Worcester Polytechnic Institute, 2020.

\bibitem{zhou2018}
Q.~Y. Zhou, J.~Park, and V.~Koltun, ``Open3d: A modern library for 3d data processing,'' {\em arXiv preprint}, 2018.

\bibitem{opencv}
{OpenCV}, ``Open source computer vision library.'' \url{https://opencv.org}, 2023.

\bibitem{numpy}
{NumPy}, ``Numpy --- official website.'' \url{https://numpy.org/}, 2024.

\bibitem{yolov10}
{YOLOv10}, ``Yolov10.'' \url{http://github.com/THU-MIG/yolov10}.

\bibitem{Roboflow_Model}
{Roboflow}, ``Polinizador dataset.'' \url{http://universe.roboflow.com/instituto-politcnico-nacional-ee9gw/polinizador}.

\bibitem{dbscan}
M.~Ester, H.~P. Kriegel, J.~Sander, and X.~Xu, ``A density-based algorithm for discovering clusters in large spatial databases with noise,'' in {\em Proceedings of the 2nd International Conference on Knowledge Discovery and Data Mining (KDD)}, pp.~226--231, 1996.

\bibitem{bellpepper}
C.~Lehnert, I.~Sa, C.~McCool, B.~Upcroft, and T.~Perez, ``Sweet pepper pose detection and grasping for automated crop harvesting,'' in {\em 2016 IEEE International Conference on Robotics and Automation (ICRA)}, pp.~2428--2434, 2016.

\bibitem{bellpepper2}
S.~Marangoz, T.~Zaenker, R.~Menon, and M.~Bennewitz, ``Fruit mapping with shape completion for autonomous crop monitoring,'' in {\em 2022 IEEE 18th International Conference on Automation Science and Engineering (CASE)}, pp.~471--476, 2022.

\bibitem{dough}
L.~A. Giefer, M.~L{\"u}tjen, A.-K. Rohde, and M.~Freitag, ``Determination of the optimal state of dough fermentation in bread production by using optical sensors and deep learning,'' {\em Applied Sciences}, vol.~9, no.~20, p.~4266, 2019.

\bibitem{trf}
M.~Branch, T.~Coleman, and Y.~Li, ``A subspace, interior, and conjugate gradient method for large-scale bound-constrained minimization problems,'' {\em SIAM Journal on Scientific Computing}, vol.~21, pp.~1--23, 1999.

\bibitem{gavin2013}
H.~P. Gavin, ``The levenberg-marquardt method for nonlinear least squares curve-fitting problems.'' \url{https://api.semanticscholar.org/CorpusID:5708656}, 2013.

\bibitem{aws}
{AWS SageMaker}, ``Aws sagemaker ground truth.'' \url{http://aws.amazon.com/sagemaker/groundtruth}.

\bibitem{ci}
J.~Ci {\em et~al.}, ``3d pose estimation of tomato peduncle nodes using deep keypoint detection and point cloud,'' {\em Biosystems Engineering}, vol.~243, pp.~57--69, 2024.

\bibitem{sun}
Q.~Sun, M.~Zhong, X.~Chai, Z.~Zeng, H.~Yin, G.~Zhou, and T.~Sun, ``Citrus pose estimation from an rgb image for automated harvesting,'' {\em Computers and Electronics in Agriculture}, vol.~211, p.~108022, 2023.

\bibitem{luo}
L.~Luo, W.~Yin, Z.~Ning, J.~Wang, H.~Wei, W.~Chen, and Q.~Lu, ``In-field pose estimation of grape clusters with combined point cloud segmentation and geometric analysis,'' {\em Computers and Electronics in Agriculture}, vol.~200, p.~107197, 2022.

\end{thebibliography}

\end{document}